\documentclass[11pt,draftcls,onecolumn]{IEEEtran}

\usepackage{amsfonts}
\usepackage{graphicx}
\usepackage{epstopdf}
\usepackage{color}
\usepackage{hyperref}

\usepackage[version=3]{mhchem}

\title{Restoring STM images via Sparse Coding: noise and artifact removal} 

\author{Jo\~ao P. Oliveira, Ana Bragan\c ca, Jos\'e Bioucas-Dias, M\'ario Figueiredo, Lu\' is Alc\'acer, Jorge Morgado, Quirina Ferreira}

\date{\today}

\begin{document}

\maketitle
\IEEEpeerreviewmaketitle

\begin{abstract}
In this article, we present a denoising algorithm to improve the interpretation and quality of scanning tunneling microscopy (STM) images. 
Given the high level of self-similarity of STM images, we propose a denoising algorithm by reformulating the true estimation problem as a sparse regression, often termed sparse coding. We introduce modifications to the algorithm to cope with the existence of artifacts, mainly dropouts, which appear in a structured way as consecutive line segments on the scanning direction. The resulting algorithm treats the artifacts as missing data, and the estimated values outperform those algorithms that substitute the outliers by a local filtering. We provide code implementations for both Matlab and Gwyddion~\cite{Necas2012}.

\end{abstract}

%
\section{Introduction}

Scanning tunneling microscopy (STM) has been considered the most important technique to image, control and monitorize  molecular systems with atomic resolution under a wide range of experimental conditions that enable molecule visualization. Typically, the STM image is composed of a bidimensional array of current values obtained when a metallic tip is very close to a conductive substrate creating a tunnel effect by applying a voltage between both conductive media or electrodes.  The image resolution  depends    on several factors,  such as the nature of electrodes, conductivity of the molecular system,  packing order of molecules,  and temperature.  The  quality of  STM images is,  however, limited by  various   degradation mechanisms including   noise (thermal and  Poissonian  due to the electronic counting process), drifts,  and dropouts.  These degradation mechanisms  reduce considerably the  interpretability  of   STM images,   calling for image denoising and  restoration  techniques to mitigate the effect of those degradations mechanisms.

There are several software suites that perform image denoise of STM images~\cite{Necas2012,soft:wsxm}. Most techniques  implement denoising in the two-dimensional frequency domain by exploiting the
fact that the STM images are quasi-periodicity and thus have  quasi-discrete and quasi-periodic   patterns in the frequency domain. However,  the degradation mechanisms mentioned above defocus the frequency  patterns, compromising the success of the frequency based  denoising techniques.

Image denoising is one of the oldest image restoration problems, which continues  to be object of intense research.  As in many  signal and image areas, the success of image denoising
depends crucially  on the existence of a good model to represent the original image. In this paper, we attack  STM denoising by reformulating the true estimation problem as a sparse regression, often termed sparse coding.  Following the standard procedure in patch-based image restoration, the image is partitioned into small overlapping square patches and the vector corresponding to each patch is modeled as a sparse linear combination of vectors, termed atoms, taken from a set called dictionary. Aiming at optimal sparse representations, and thus at optimal noise removing capabilities, the dictionary is learned from the data it represents via matrix factorization with sparsity constraints on the code (i.e., the regression coefficients) enforced by the $\ell_1$ norm~\cite{art:EladImageDenoising,art:MairalSparseColorRestoration}.

The sparsity of a representation on a dictionary  is strongly linked with the level of  self-similarity of the images under study. It happens that  STM images   have a high level of self-similarity and, therefore,  they are natural candidates to be denoised via sparse regression on learned  dictionaries. Compared with the standard approaches, STM images pose an additional challenge: they have plenty of artifacts, mainly  dropouts, which appear in a structured way as consecutive line segments  on the scanning direction.  To cope we these artifacts,  we  modify the algorithm introduced in
\cite{art:EladImageDenoising,art:MairalSparseColorRestoration};  the modifications consists in  treating the artifacts as missing data in both the dictionary learning and denoising  (i.e., coding) phases.

To assess the quality of the proposed approach, we applied the algorithm to STM images of porphyrins and bipyridines adsorbed on highly oriented pirolitic graphite. The Matlab code of the algorithm and a Gwyddion module can be obtained from  \href{http://www.lx.it.pt/~jpaos/stm/}{http://www.lx.it.pt/$\sim$jpaos/stm/}.

%
\section{Sparse Image representation}

Signals and images can be represented in different ways. Several methods include Fourier Transform methods~\cite{bookFourier2D}, wavelets methods~\cite{mallat}, curvelets methods~\cite{art:curvelets}, just to mention a few.
Despite the underlying differences, each representation has its own advantages when  used in some of the classical, and still challenging, image restoration problems~\cite{book:Jain,book:handbook:bovik}.

Recently, sparse linear decomposition of signals  in learned dictionaries led to state-of-the-art methods. The idea is to represent a signal, or fragments of the signal denoted patches, as a linear combination of vectors, termed atoms, taken from a set called dictionary. The dictionary can be fixed, for example based on wavelets~\cite{mallat}, or can be learned~\cite{art:ksvd}.

The topic of sparse and redundant representations of real world images is currently one of the most active research fields. Recently, there has been an increased interest due to the fact that dictionaries yielding sparse representations may be learned directly from the data they represent~\cite{art:EladImageDenoising,art:MairalSparseColorRestoration}.

The fact that the real world images (and signals) admit sparse representations in suitable dictionaries is a consequence of the high level of self-similarity of real world images (and signals), i.e., given an image patch, there is a high likelihood of finding similar patches at different locations and scales. This is also true in the case of STM images: by nature, an STM image is almost a periodic image, and thus there exist similar patches with high probability.

The proposed algorithm works as follows: (i) the noisy STM image is decomposed in smaller patches; (ii) a dictionary is learned from the data; (iii) the patches are sparsely coding using just a few atoms from the learned dictionary; and (iv) the denoised/restored image is obtained by composition of the coded patches.
In the following sections we introduce notation and describe in detail the constituent parts of the proposed \textit{sparseland} framework to restore STM images.

\subsection{Patch decomposition and composition of an image}

Sensor measurements are usually corrupted by noise. STM images are not an exception, and
the  observation mechanism is well approximated by the additive model
\[
\mathbf{z}\,=\,\mathbf{x}+\mathbf{n},
\]
where $\mathbf{z}$ is the noisy image, $\mathbf{x}$ the true image, and $\mathbf{n}$ models noise and model errors.
The  noise is originated by  different sources, such as the tip movements, thermal noise, Poissonian couting noise, or even
from the fact that different molecules have different conductivities. Given the relatively number of independent causes,
we  assume that the noise is  zero-mean  independent  Gaussian  with an unknown variance $\sigma$.
The goal is to obtain an estimate of $\mathbf{x}$  given $\mathbf{z}$.

Unlike other denoise algorithms, the proposed \textit{sparseland} framework processes the image in small patches rather than the whole image. Consider the noisy image $\mathbf{z}\,\in\,\mathbb{R}^{N}$, where $N\,=\,N_1\times N_2$ is the image size, and the patch $\mathbf{z}_i\,\in\,\mathbb{R}^m$ containing the pixels located inside a square window of size $\sqrt{m} \times \sqrt{m}$ centered at the $i$-th pixel. The total number of overlapping patches that exist in image $\mathbf{z}$ is $N_p = (N_1 - \sqrt{m} + 1)(N_2 - \sqrt{m} + 1)$.
Let $\mathbf{x}_i\,\in\,\mathbb{R}^m$ and $\mathbf{n}_i\,\in\,\mathbb{R}^m$ denote two vectors holding, respectively, the elements $\mathbf{x}$ and of $\mathbf{n}$ corresponding to the $i$-th patch. We have then
$$\mathbf{z}_i\,=\,\mathbf{x}_i+\mathbf{n}_i,\quad i\,=\,1,...,N_p.$$
In the \textit{sparseland} framework, patches are processed individually, and thus for each $\mathbf{x}_i$ we obtain an estimate $\widehat{\mathbf{x}}_i$. Given that the reconstruction is not exact, we can write
\[
\widehat{\mathbf{x}}_i\, =\, \mathbf{x}_i + \varepsilon_i, \quad i\,=\,1,...,N_p,
\]
where $\varepsilon_i$ is the estimation error for the $i$-th patch.
To produce the overall estimate of $\mathbf{x}$ from the estimates $\mathbf{x}_i$, for $i\,=\,1, \ldots, N_p$, we combine the information of several patches as follows.
We introduce selection matrices $\mathbf{S}_i$ such that $\mathbf{x}_i\,=\,\mathbf{S}_i\,\mathbf{x}$. Note that each row of $\mathbf{S}_i$ contains just one non-null element of value 1. Consider also  the following matrix and vectors:
\begin{eqnarray*}
\mathbf{S} &:= &[\mathbf{S}_1^T, \ldots, \mathbf{S}_{N_p}^T]^T\\
\widehat{\mathbf{x}}_P &:= &[\widehat{\mathbf{x}}_1^T, \ldots, \widehat{\mathbf{x}}_{N_p}^T]^T\\
\varepsilon_P &:= &[\varepsilon_1^T, \ldots, \varepsilon_{N_p}^T]^T.
\end{eqnarray*}
With the above notation, the estimate of each patch  is given by
\[
\widehat{\mathbf{x}}_P\,=\,\mathbf{S}\,\mathbf{x} + \varepsilon_P,
\]
and the overall estimate $\widehat{\mathbf{x}}$ can be obtained by
\begin{equation}
\label{eq:estimate_x_from_xp}
\widehat{\mathbf{x}}\,=\,\mathbf{S}^\dagger\,\widehat{\mathbf{x}}_P,
\end{equation}
where $\mathbf{S}^\dagger\,:=\,(\mathbf{S}^T\mathbf{S})^{-1}\,\mathbf{S}^T$.
Noting that $\mathbf{S}^T\,\mathbf{x}_i$ places the path number $i$ at its position in the image and that $(\mathbf{S}^T\mathbf{S})$ is a diagonal matrix whose $i$-th diagonal element holds the number of times pixel $i$ appears in any patch, the estimate for the $i$-th pixel is the average of all its estimates, one per patch containing thereof. This condition implies that $(\mathbf{S}^T\mathbf{S})$ is invertible as far as any pixel belongs at least to one patch.

The estimate given by~(\ref{eq:estimate_x_from_xp}) is the best linear unbiased estimator for $\mathbf{x}$~\cite{book:kay}, as long as we assume that $\varepsilon_P$ is zero-mean and independent and identicaly distributed (iid), and thus having a covariance matrix proportional to the identity matrix, i.e., $\mathbf{C}_\varepsilon\,\approx\,\mathbf{I}$. However, due to the overlapping structure of the patches it is very difficult to accurately compute $\mathbf{C}_\varepsilon$. We opted to use the suboptimal estimate~(\ref{eq:estimate_x_from_xp}) that, nevertheless, produces state-of-the-art results.

\subsection{Dictionary Learning}
\label{sec:diclearning}

Given the periodic characteristic of a STM image, it is very likely that it admits a sparse representation in fixed existing dictionaries (i.e., representations). Wavelets and Fourier
bases are two paradigmatic examples. However,  the use of learned dictionaries  from the  data they represent holds systematically the state-of-the-art~\cite{art:EladImageDenoising,Mairal2009}.
The goal of dictionary learning is to find a dictionary such that the patches can be accurately represented with just a few number of  elements. One way to formulate this idea is solving the following optimization problem:
\begin{equation}
\label{eq:diclearning}
\min_{\mathbf{D}\in C, \alpha_1, \ldots, \alpha_{N_P}} \sum_{i=1}^{N_p}(1/2)\|\mathbf{z}_i-\mathbf{D}\mathbf{\alpha}_i  \|_2^2 + \lambda\|\mathbf{\alpha}_i \|_1,
\end{equation}
where $\mathbf{D} \equiv [\mathbf{d}_1, \ldots, \mathbf{d}_k] \in \mathbb{C}^{m\times k}$, $C\,:=\,\{ \mathbf{D} \in \mathbb{C}^{m\times k} : | \mathbf{d}_j^T\mathbf{d}_j| \leq 1,\; j=1, \ldots, k  \}$. The quadratic terms account for the representation errors and the $l_1$ norm is used to promote sparse codes. The relative weight between the two terms is established by the regularization parameter $\lambda > 0$. The constraint $\mathbf{D} \in C$ is just a technical condition to prevent $\mathbf{D}$ from being arbitrarily large.

Although the focus of this paper is not on the optimization of (\ref{eq:diclearning}), it is worth mentioning just a few words about it. A common approach to handle these problems~\cite{Lee07efficientsparse,art:ksvd,Olshausen97sparsecoding,Rakotomamonjy2013126} is to alternate the optimization with respect to $\mathbf{D}$ and $\alpha_1, \ldots, \alpha_{N_P}$. The optimization with respect to $\mathbf{D}$ is a quadratic problem with convex constraints and optimization with respect to $\alpha_1, \ldots, \alpha_{N_P}$ is also convex and decoupled (we can solve individually for each $\alpha_i$ variable). However, the optimization with respect to all variables is non-convex. Although we are not guaranteed to converge to the global minimum, the obtained stationary points have shown to produce state-of-the-art results.

In a typical STM image, we have $N_p\, =\, 60000$, $m = 100$, and $k = 64$. These numbers  mean that the  computational burden necessary to compute (\ref{eq:diclearning}) is heavy.
To cope with this shortcoming, we adopt the online learning approach introduced in~\cite{Mairal2009}, which largely reduces that computational complexity. For more details see  in~\cite{art:bioucasComplexSparse}.

\subsection{Sparse Coding}
\label{sec:sparsecoding}
Once we have learned a dictionary $\mathbf{D}$ with respect to which the patches of $\mathbf{x}$ admit a sparse representation, we proceed and compute the estimate of a given patch $\mathbf{x}_i\,=\,\mathbf{S}_i\mathbf{x}$, for $i\in \{1,\ldots,N_p\}$.
Following the standard formulation in synthesis based approaches to sparse regression, the patch estimate $\widehat{\mathbf{x}}_i\,=\, \mathbf{D} \widehat{\alpha}$ is the solution of the constrained optimization
\begin{equation}
\label{eq:sparsedenoise}
\min_{\mathbf{\alpha}} \| \mathbf{\alpha}\|_0\quad \text{subject to:}\; \| \mathbf{D}\mathbf{\alpha} - \mathbf{z}_i \|^2 \leq \delta,
\end{equation}
where $\mathbf{z}_i\,=\,\mathbf{S}_i\mathbf{z}$ is the observed patch corresponding to the true patch $\mathbf{x}_i$, $\alpha \in \mathbb{C}^k$ and $\| \mathbf{\alpha}\|_0$ is the number of nonzero elements of vector $\mathbf{\alpha}$, and $\delta \ge 0$ is a parameter controlling the reconstruction error.

When $\delta\,=\,0$ and the dictionary is full column range, the solution of the optimization problem~(\ref{eq:sparsedenoise}), if it exists, is easy to compute. However, in the majority of the situations, this is not the case because the dictionaries are often overcomplete and,  due to noise, $\delta > 0$. Under these circumstances, the problem is NP-hard~\cite{Natarajan:1995:SAS:207985.207987}. Two different approaches have been followed. One consists in replacing $\|\cdot \|_0$ by the $l_1$ norm, which results in the so-called least absolute shrinkage and selection operator (LASSO)~\cite{Tibshirani94regressionshrinkage}, which is equivalent to the basis pursuit denoising (BPDN)~\cite{Chen98atomicdecomposition}. The other consists in attacking directly the original problem using greedy algorithms such as the orthogonal basis pursuit (OMP)~\cite{Pati93orthogonalmatching}, iterative hard thresholding (IHT)~\cite{Blumensath_iterativehard}, hard thresholding pursuit~\cite{Foucart_hardthresholding}, or approximate message passing (MP)~\cite{Vila11expectation-maximizationbernoulli-gaussian}.

In this work we follow the same approach as in~\cite{art:bioucasComplexSparse}, and choose the OMP~\cite{Pati93orthogonalmatching} algorithm because of its lower computational complexity compared with the others. The details can be found in~\cite{art:bioucasComplexSparse}.

%
\section{Noise and Artifact removal}

Assuming that the noise of each patch is zero-mean and has covariance $\sigma^2\mathbf{I}$, it is possible to show~\cite{art:bioucasComplexSparse} that the relative attenuation of noise is given by
\begin{equation}
\label{eq:noisereduction}
\frac{\mathbb{E}[\|\varepsilon_i \|^2_2]}{\mathbb{E}[\|n_i \|^2_2]} = \frac{p}{m}.
\end{equation}
This means that the estimation error is proportional to the sparsity level of the signal. However, in practice,
the ratio~(\ref{eq:noisereduction}) is hardly achieved. One can also understand the noise reduction from a geometric point of view. The sparse coding of the patches, given by optimization problem (\ref{eq:sparsedenoise}), can be viewed as a projection of the noisy patch $\mathbf{z}_i$ onto the subspace spanned by the active columns of $\bf D$ in that patch, corresponding to the non-zero components of $\mathbf{\alpha}$.  Given that the number of active actoms is much samller than the  $m$, the noise is largely reduced.

\begin{figure}
 \includegraphics[width=7cm]{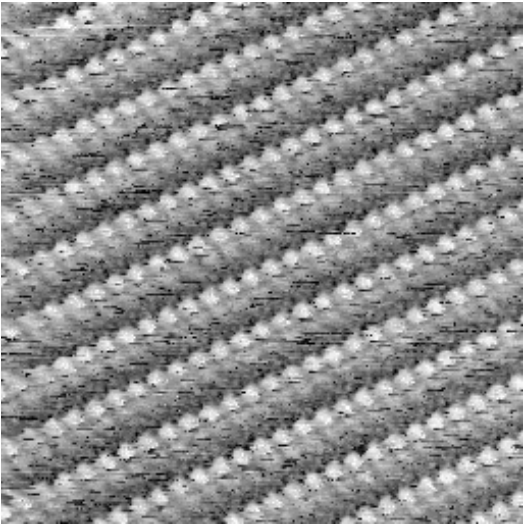}%
 \caption{STM image (30$\times$30 nm, V=570mV, I=28pA) of a polymorphic phase of a ZnOEP monolayer absorbed on  HOPG where the ZnOEP are organized as dimers with different brightness. The image has important artifacts that make it hard to interpret.\label{fig:artifacts}}%
\end{figure}

The \textit{sparseland} framework works well for STM images because the images are almost periodic. This makes it easier to learn the dictionary directly from the data. Besides the presence of noise, STM images usually have plenty of artifacts. However, they do not appear isolated, such as impulsive noise, but rather in a structured way. Typically, several consecutive pixels along a scanning line may be corrupted. Figure~\ref{fig:artifacts} shows an example of a STM image with artifacts that hampers the observation of darker spots. The image represents a polymorphic monolayer with zinc(II)octaethylporphyrin (ZnOEP) molecules organized as dimer-like units with different brightness aligned along parallel rows and absorbed on highly oriented pyrolitic graphite (HOPG) by $\pi-\pi$ stacking interaction~\cite{art:quirina1}. The difference in brightness of the two ZnOEP units is due to their different coupling to the HOPG where the darker molecules have lower signal and for this reason is hard to distinguish with high detail the central part of porphyrines and their axial groups.

In order to proceed, special care must be taken to eliminate these artifacts, otherwise the learned dictionary  ends up with patches that can represent well these outliers.
To overcome these problems, we change the algorithm by introducing an extra vector that masks out these pixels. This approach has several advantages. It makes it easy to iteratively improve on current results, either by adding or removing pixels from the mask (considered outliers). It also allows us to easily fill in the outliers pixels (inpainting) with new values. As opposed to a classical local filtering method, the new inpainting values can take into account the full image statistics, and thus attain superior reconstruction quality.

\subsection{Proposed algorithm}

\begin{figure}
 \includegraphics[width=7cm]{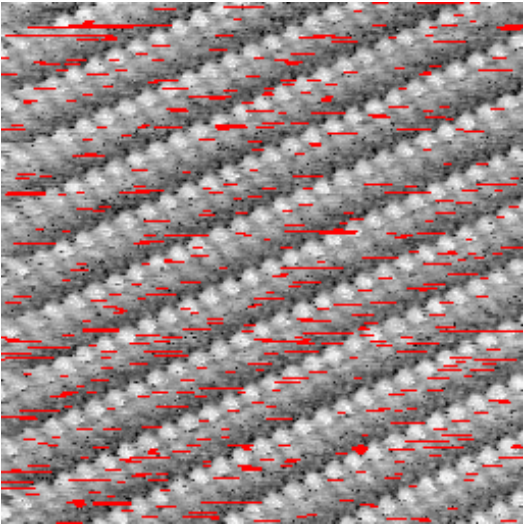}%
 \caption{STM image of Figure~\ref{fig:artifacts} with masked outliers represented in red.\label{fig:mask}}%
\end{figure}

The proposed algorithm changes the previously introduced \textit{sparseland} framework to accomplish two objectives: (i) deal with the structured outliers; and (ii) do inpainting of those pixels. We start by defining a binary vector $\mathbf{m} \in \{0,1\}^N$ of size $N = N_1 \times N_2$, where 0 indicates that the corresponding pixel in $\mathbf{x}$ is considered an outlier. The purpose of this vector is to mask out the outliers.
Let  $\mathbf{m}_i = \mathbf{S}_i\,\mathbf{m}$ for $i\in \{1,\ldots,N_p\}$ denote a patch of $\bf m$. Note that $\mathbf{m}_i$ indicate the outlier's positions in the patches $\mathbf{x}_i$. Figure~\ref{fig:mask}  shows the STM image of Figure~\ref{fig:artifacts} with the mask over-imposed. The pixels considered outliers are represented in red.

The dictionary learning algorithm is the same as described in section~\ref{sec:diclearning}. To avoid learning the outliers structures, instead of considering all the patches, we use $\mathbf{m}_i$ to discard those that contain outliers.
The sparse coding algorithm is also the same as described in section~\ref{sec:sparsecoding}.
However, in this case, we do not discard any patch; we code the patches ignoring only the pixels that are considered outliers by solving the  optimization 
\begin{equation}
\label{eq:sparsedenoiseproj}
\min_{\mathbf{\alpha}} \| \mathbf{\alpha}\|_0\quad \text{subject to:}\; \| \mathbf{m}_i \odot(\mathbf{D}\mathbf{\alpha} - \mathbf{z}_i) \|^2 \leq \delta,
\end{equation}
where the symbol $\odot$ represents elementwise multiplication.
 Note that the effect of $\mathbf{m}_i\odot\bf  D$ and $\mathbf{m}_i\odot{\bf z}_i$ is  to  project the columns of $\bf D$ and ${\bf z}_i$ onto a lower dimension subspace. The corresponding change in the OMP algorithm is simply an element-wise multiplication by~$\mathbf{m}_i$.

Finally, once we have a sparse representation for all the patches, the image composition is the pixel-wise average of all the patches covering a pixel, marked or not as an outlier. The inpainting values obtained outperform those that result from substituting the outliers by a local filtering. In section~\ref{sec:results} we compare both approaches. A Matlab version and a Gwyddion\cite{Necas2012} module of the code are available at  \href{http://www.lx.it.pt/~jpaos/stm/}{http://www.lx.it.pt/$\sim$jpaos/stm/}.

\subsection{Pre-processing and artifact selection}

Before we apply the proposed algorithm, raw STM images are pre-processed and artifacts are identified, i.e., matrix $\mathbf{m}$ is defined. We start by applying the following pre-processing steps:
\begin{enumerate}
\item mean level subtraction: a plane is adjusted to all the data points and subtracted from the data;
\item outlier removal: pixel data that is below 1\% or above 99\% of the quantile are removed;
\item leveling by median subtraction: lines are adjusted to match height median.
\end{enumerate}

Following this pre-processing steps, artifacts are identified by inspection and marked in matrix $\mathbf{m}$. Remark that common image processing software suites have tools designed to obtain $\mathbf{m}$ automatically. As an example, Gwyddion~\cite{Necas2012} implements an algorithm denoted \textit{Mark Scars}, where several parameters for outlier detection can be specified such as minimum/maximum length and width.

For better results, the process can be iterated: start with an initial mask, apply the algorithm, improve on the mask from restored image and repeat the process.

\begin{figure*}
\begin{tabular}{ccccc}
\includegraphics[width=3.0cm]{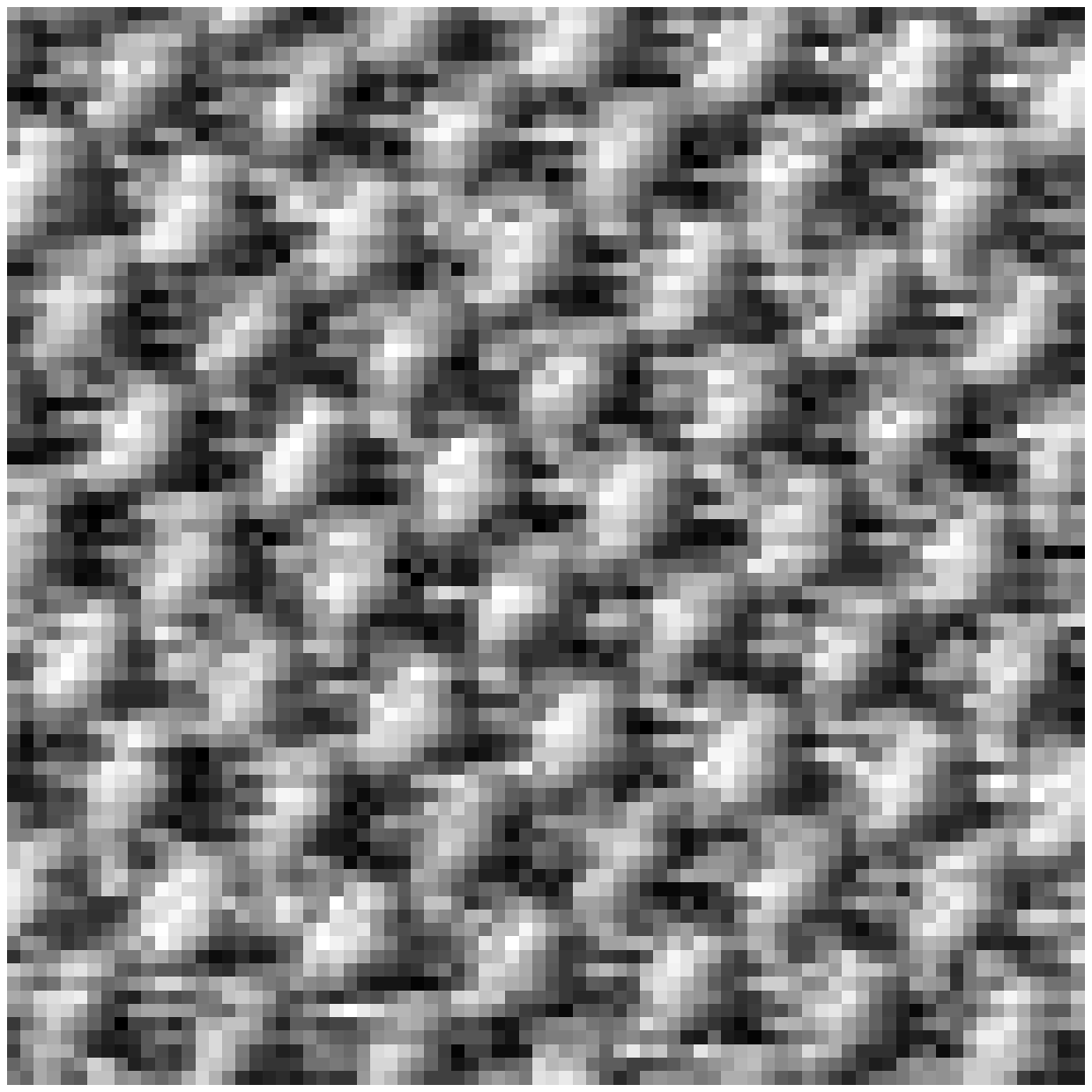} &
\includegraphics[width=3.0cm]{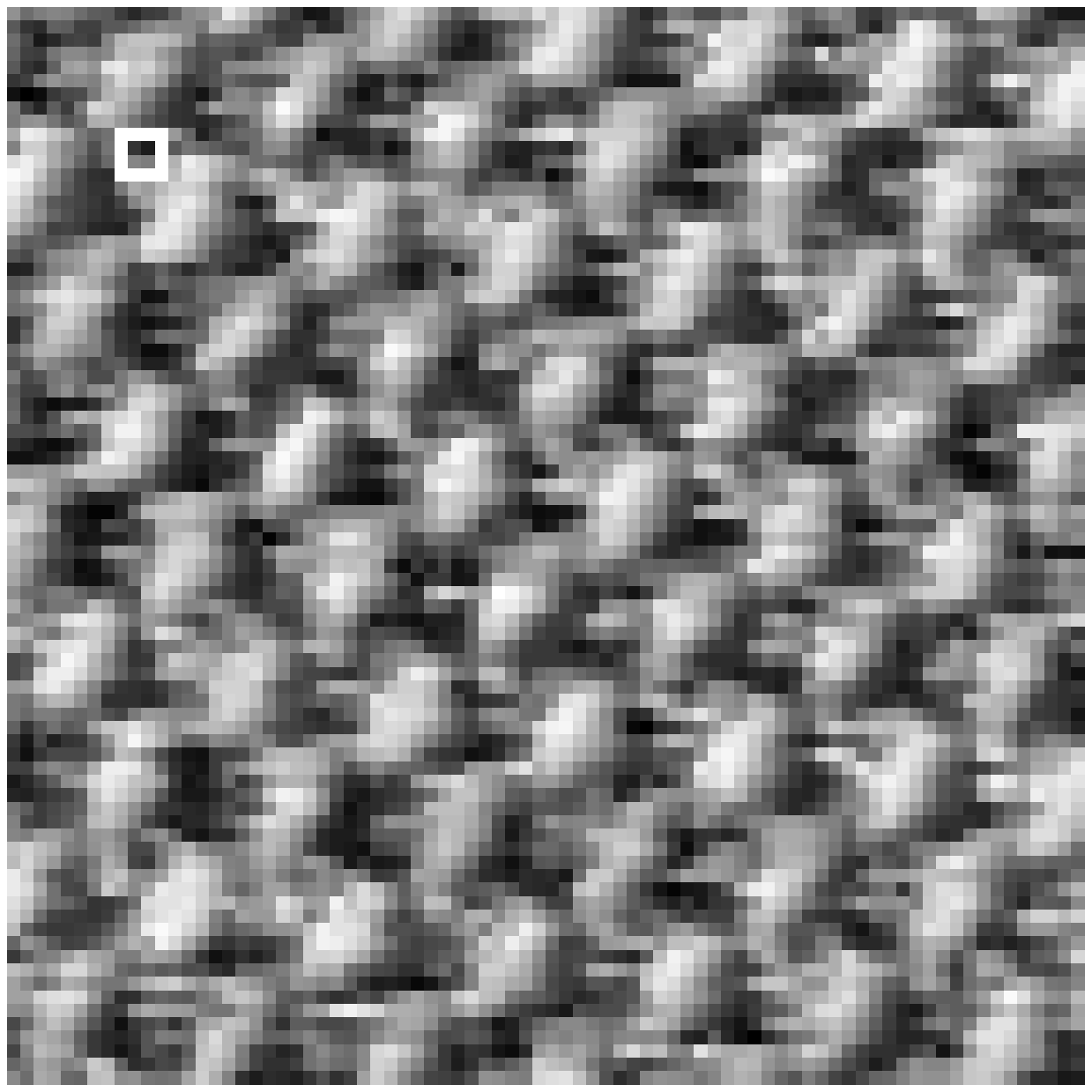} &
\includegraphics[width=3.0cm]{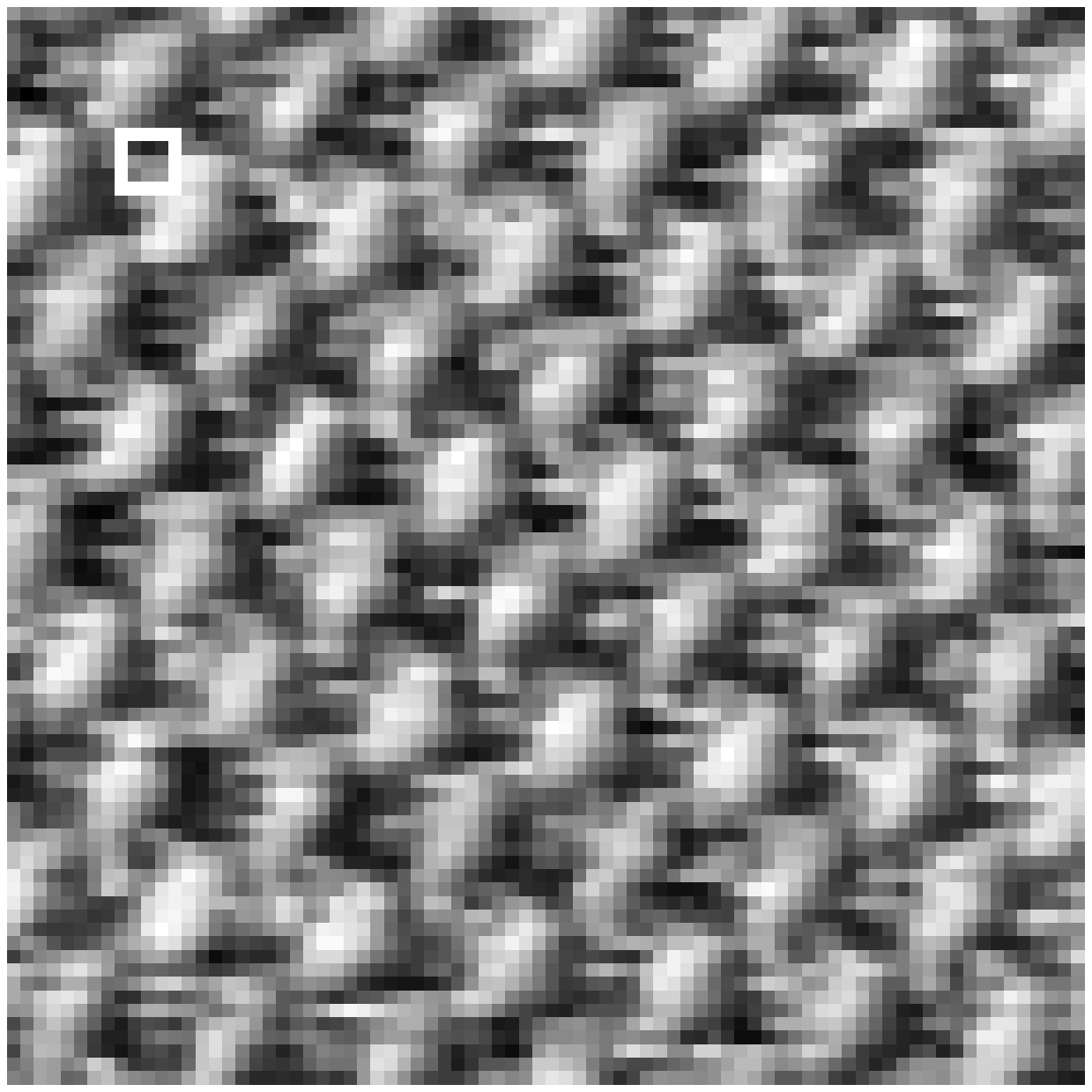} &
\includegraphics[width=3.0cm]{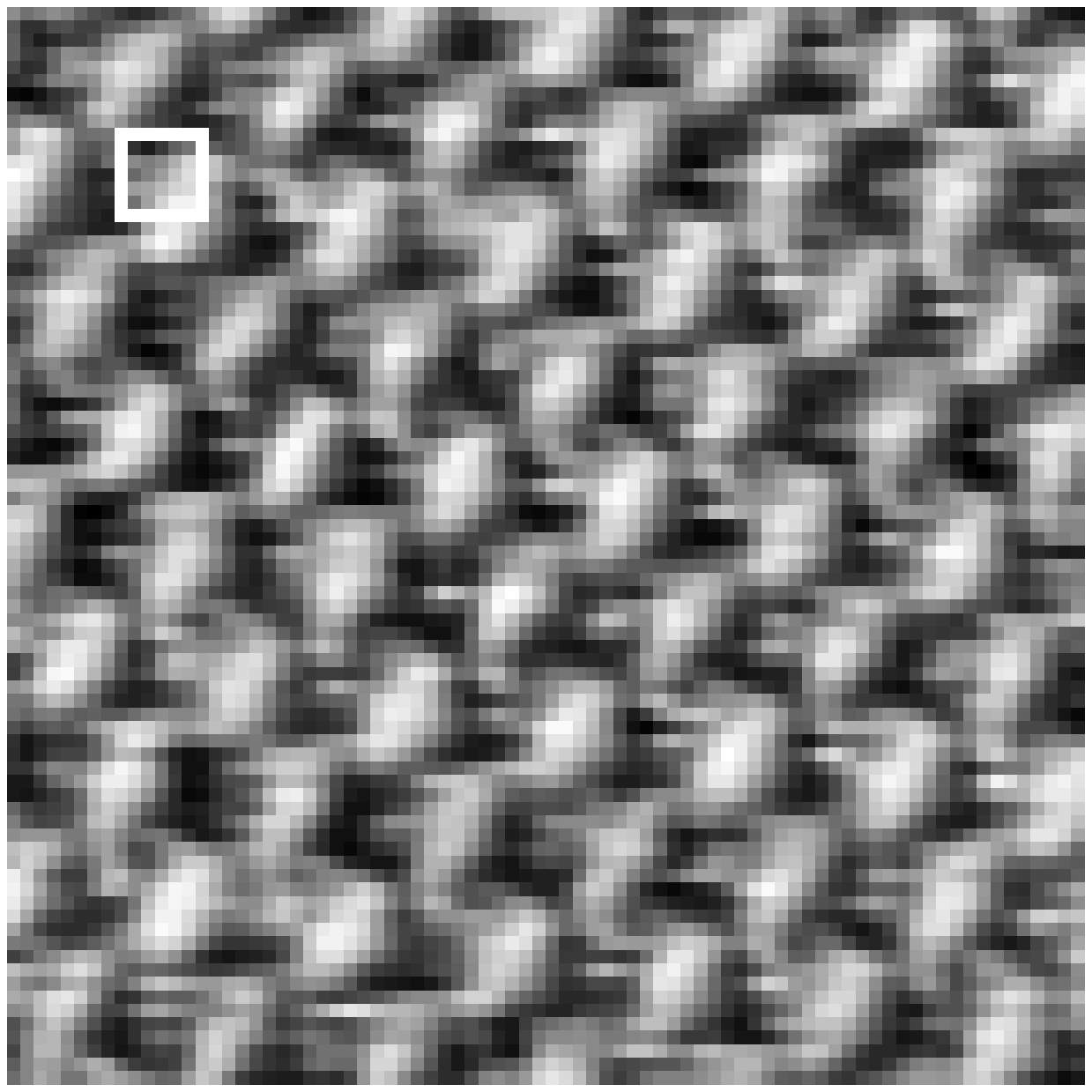} &
 \includegraphics[width=3.0cm]{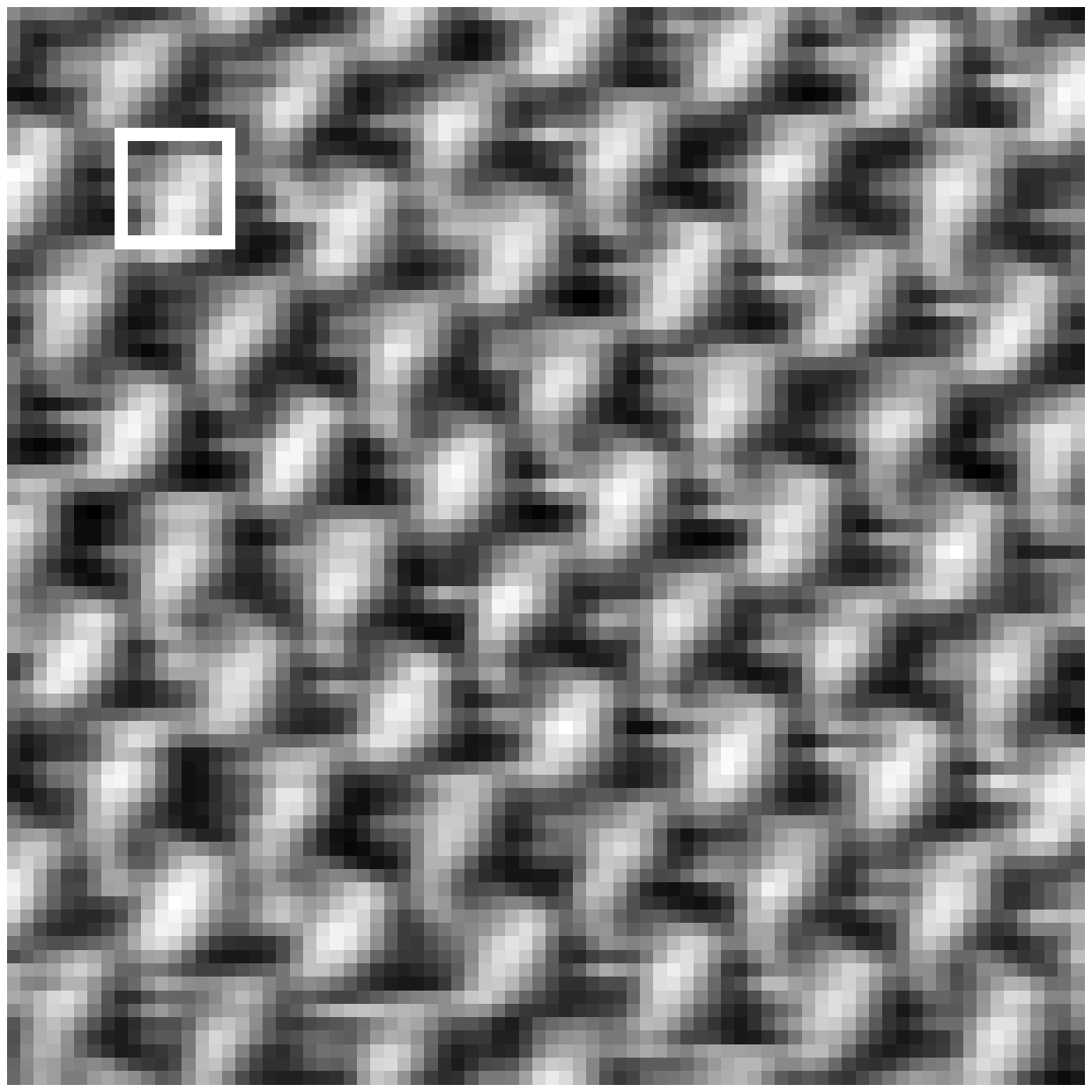} \\*[-0pt]
  (a) noisy & (b) $\sqrt{m}\,=\,3$ &
  (c) $\sqrt{m}\,=\,4$ & (d) $\sqrt{m}\,=\,6$ & (e) $\sqrt{m}\,=\,8$\\*[4pt]
\includegraphics[width=3.00cm]{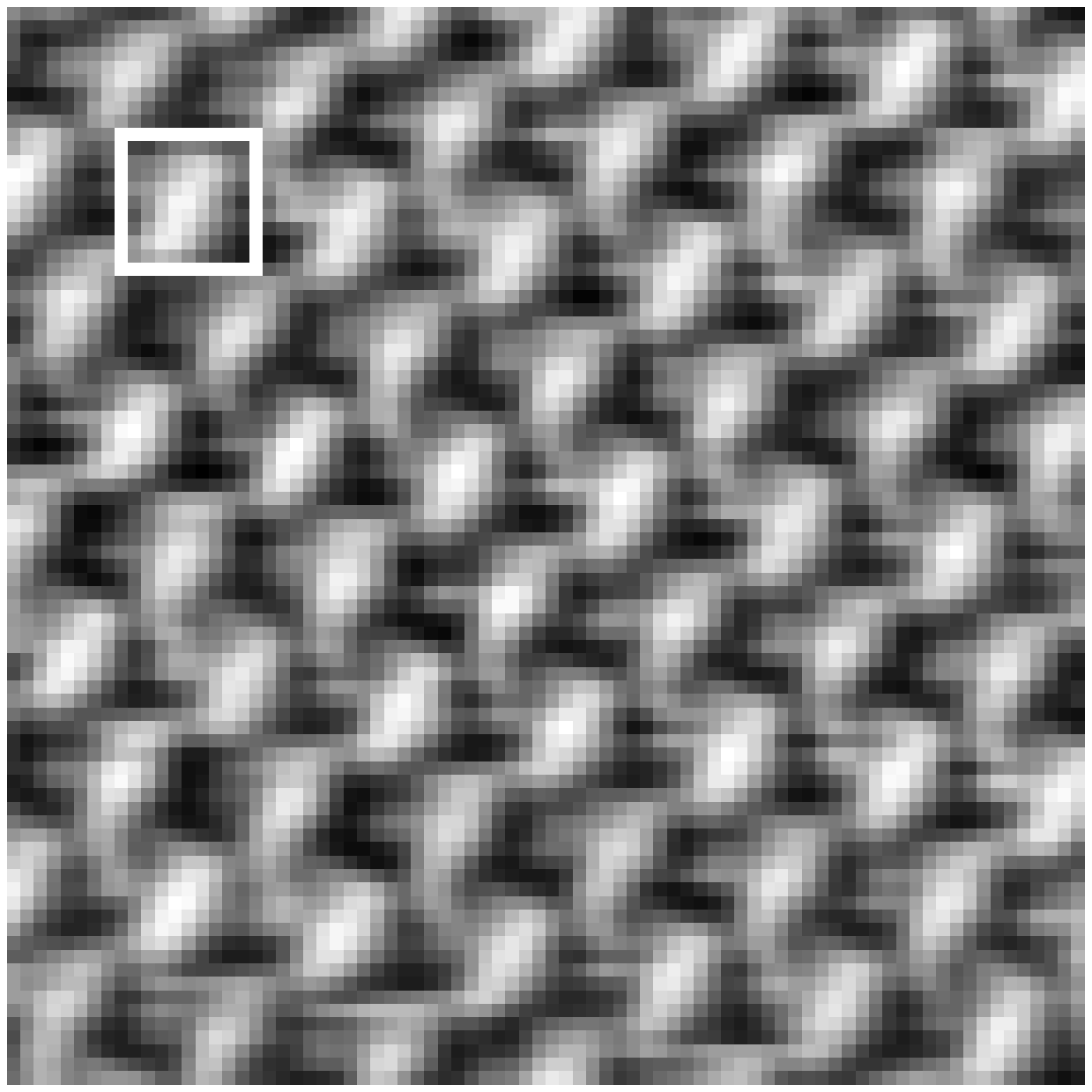} &
\includegraphics[width=3.0cm]{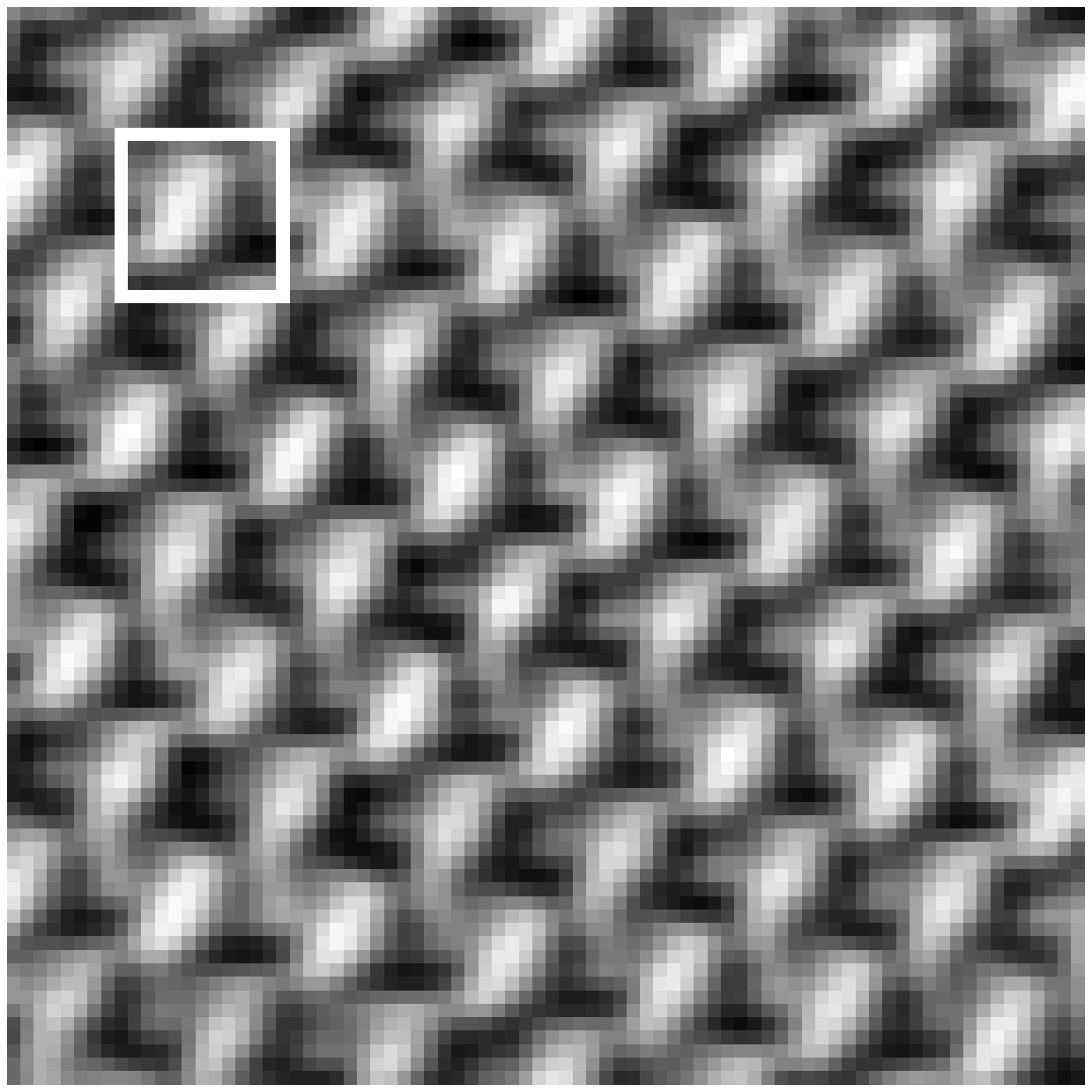} &
\includegraphics[width=3.0cm]{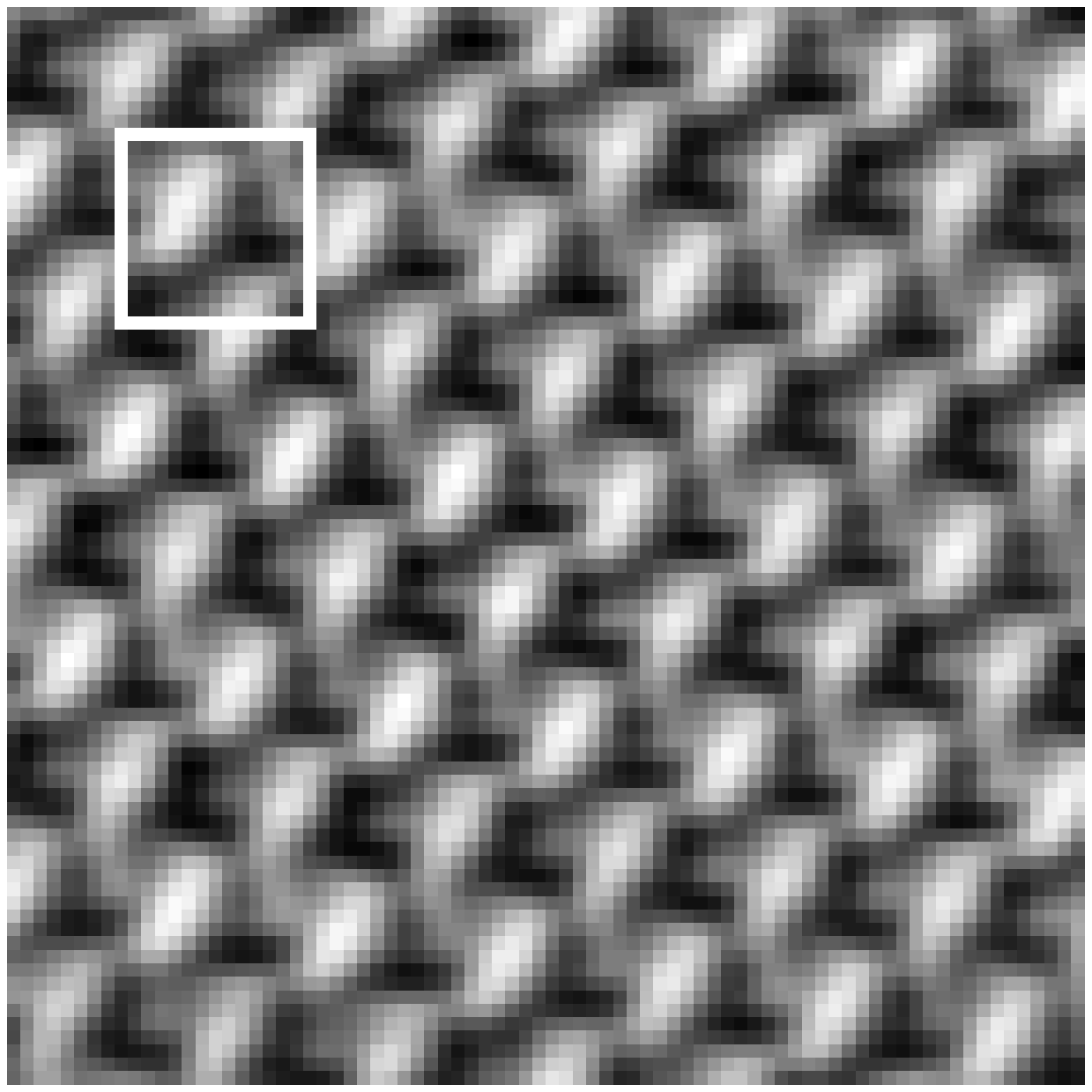} &
\includegraphics[width=3.0cm]{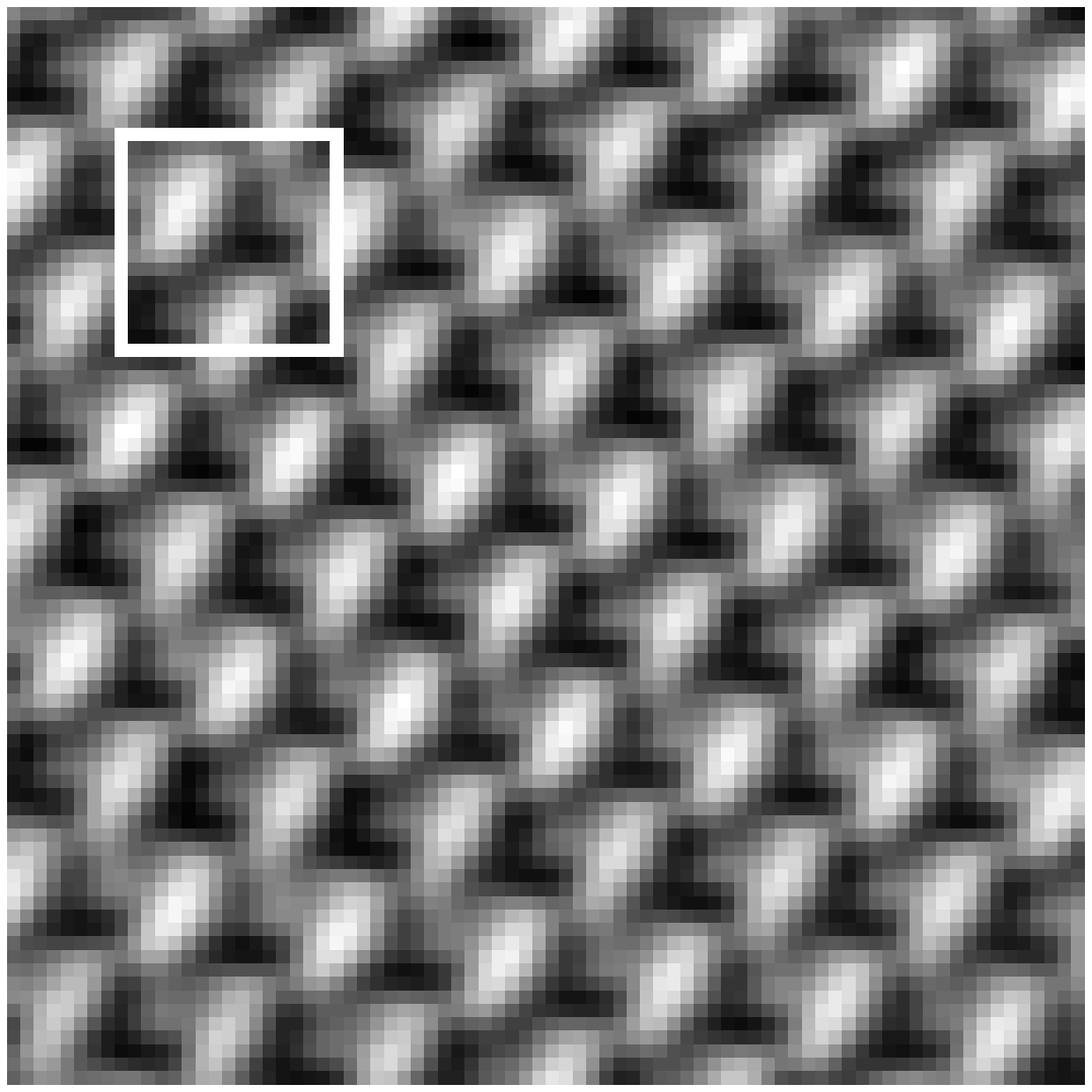} &
\includegraphics[width=3.0cm]{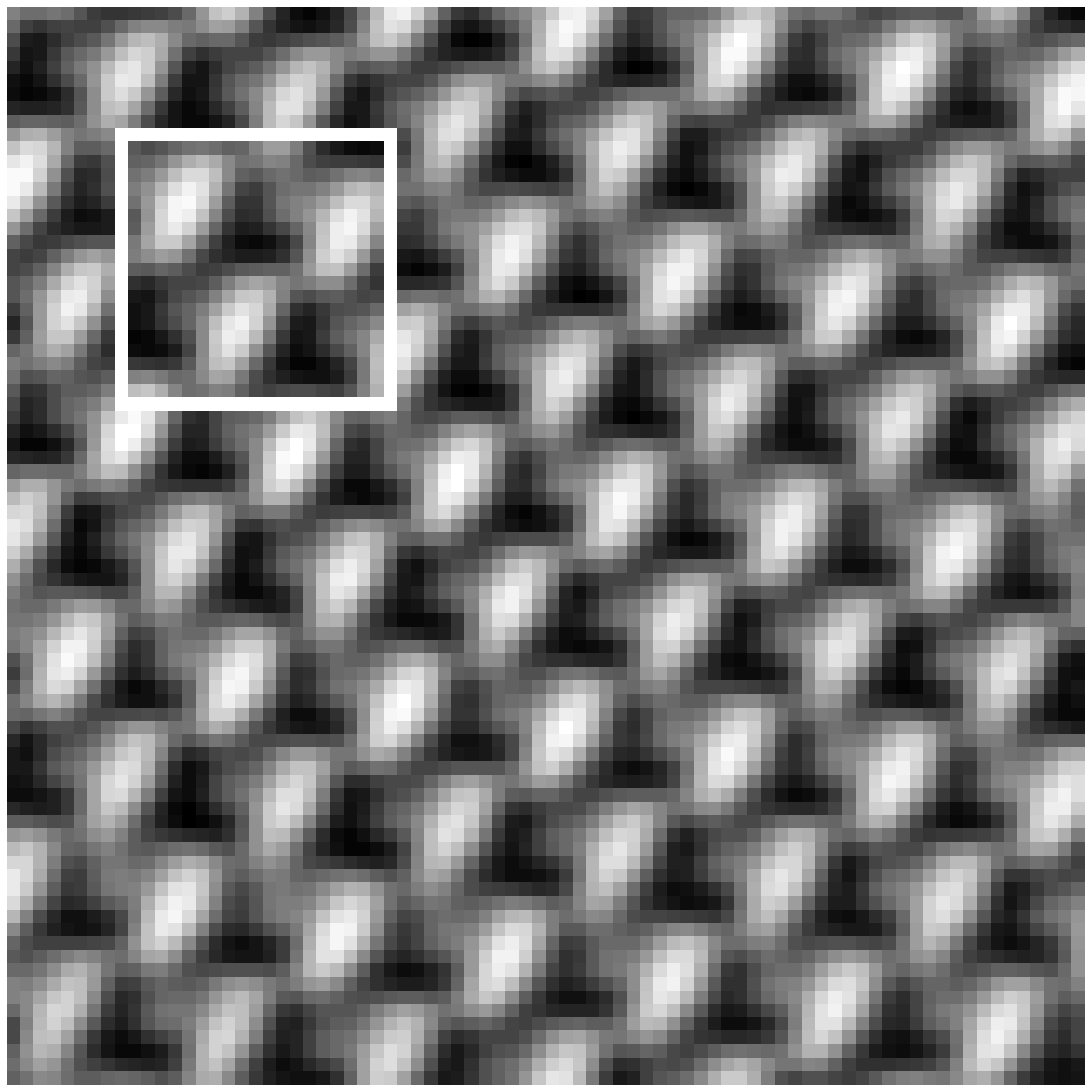} \\*[-0pt]
 (f) $\sqrt{m}\,=\,10$ &  (g) $\sqrt{m}\,=\,12$ & (h) $\sqrt{m}\,=\,14$ &
 (i)  $\sqrt{m}\,=\,16$ & (j) $\sqrt{m}\,=\,20$\\*[-2pt]
\end{tabular} 
 \caption{(a) STM image (10$\times$10nm, V=0.56V, I=0.2nA) of an incomplete BP monolayer absorbed on a full packed monolayer of ZnOEP that is absorbed on HOPG substrate. Each bright region correspond to a molecular wire formed by HOPG/ZnOEP/BP alternated by unreacted ZnOEP molecules (darker domains); (b)--(j) close up of STM image denoise with different patch sizes. The small white square represents the patch size at the image's scale.\label{fig:stm1patches}}%
\end{figure*}

\subsection{Parameter selection}

There are several parameters that affect the behavior of the overall algorithm. In this section we discuss how these parameters can be tuned in order to obtain the best possible outcome.
The algorithm is made up of two parts: dictionary learning and sparse coding.

The first parameter that we need to define is the patch size, that is common for both parts.
Patch size influences the amount of noise that is possible to remove. Bigger patch sizes results in  better noise reduction.
However, the patch size influences how data can be represented by the dictionary. Smaller patch sizes means that the dictionary adapts better to data, including details and noise, while larger patch sizes capture the similarities and periodic components.

\begin{figure}
\includegraphics[width=8.5cm]{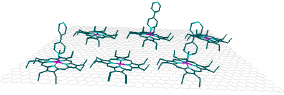}
\caption{Molecular schema illustrating the HOPG/ZnOEP/BP molecular wires alternated by ZnOEP absorbed on HOPG.\label{fig:stm1schema}}%
\end{figure}

From the exhaustive experiments, we concluded that the patch size should be at least equal to the smallest interest object that can be represented by $\sqrt{m}\times \sqrt{m}$ pixels.
Figure~\ref{fig:stm1patches}~(a) shows an incomplete monolayer with 4,4'-bipyridine (BP) absorbed on top of a full  packed monolayer of ZnOEP through a Zn-N bond. Each pair of ZnOEP on BP forms a vertical molecular wire absorbed on a HOPG substrate~\cite{art:quirina2}. The brighter regions are attributed to the BP molecules bonded to Zn alternated with darker domains that correspond to unreacted ZnOEP molecules. The corresponding molecular schema is depicted in Figure~\ref{fig:stm1schema}.

Figures~\ref{fig:stm1patches}~(b)--(k) show the results obtained by the proposed algorithm by varying the patch size. As it can be seen, smaller patch sizes adapt better to the noise image, and thus attain smaller noise reduction. On the other hand, larger patch sizes result in an almost periodic image, where important details are lost.
For this particular image --- Figure~\ref{fig:stm1patches} ---, the best results are obtained when the patch size is around $\sqrt{m}=10$.
In all experiments the regularization parameter was set at $\lambda = 0.11$, and the dictionary consisted of 64 atoms.

For the sparse coding algorithm (\ref{eq:sparsedenoiseproj}), two  parameters need to be set: the maximum number of atoms (MNA) and the error parameter~$\delta$. These parameters determine the reconstruction \textit{error} of patch $\mathbf{z}_i$: although 1 or 2 patches aren't enough to correctly represent data, a larger number gives more flexibility and adaptability, and thus less noise reduction.
The implementation of algorithm~\cite{art:bioucasComplexSparse} searches for the sparsest number of atoms that attain an error less than $\delta$, with at most the MNA. Note that the MNA is a hard threshold, i.e., each patch is not reconstructed with more atoms than that.

The error parameter~$\delta$ depends on the noise statistics of the image.
Based on a Gaussian noise distribution assumption, an estimate~\cite{art:bioucasComplexSparse} for $\delta$ is given by $\delta = \sigma^2\, F^{-1}_{\chi^2(m)}(\gamma)$, where the noise variance $\sigma$ is estimated by the MAD rule~\cite{Donoho94idealspatial}. In all the experiments we set $\gamma = 0.96$.

\section{Results}
\label{sec:results}

In this section we assess the merit of the proposed algorithm applied on the STM image of polimorphic phase of a ZnOEP monolayer absorbed on HOPG, previously depicted in Figure~\ref{fig:artifacts}. We selected for the patch size $\sqrt{m} = 10$, the maximum number of atoms was set at 4 and $\delta = 10$. The mask used is depicted in Figure~\ref{fig:mask}.

\begin{figure}
\begin{tabular}{cc}
 \includegraphics[width=3.4cm]{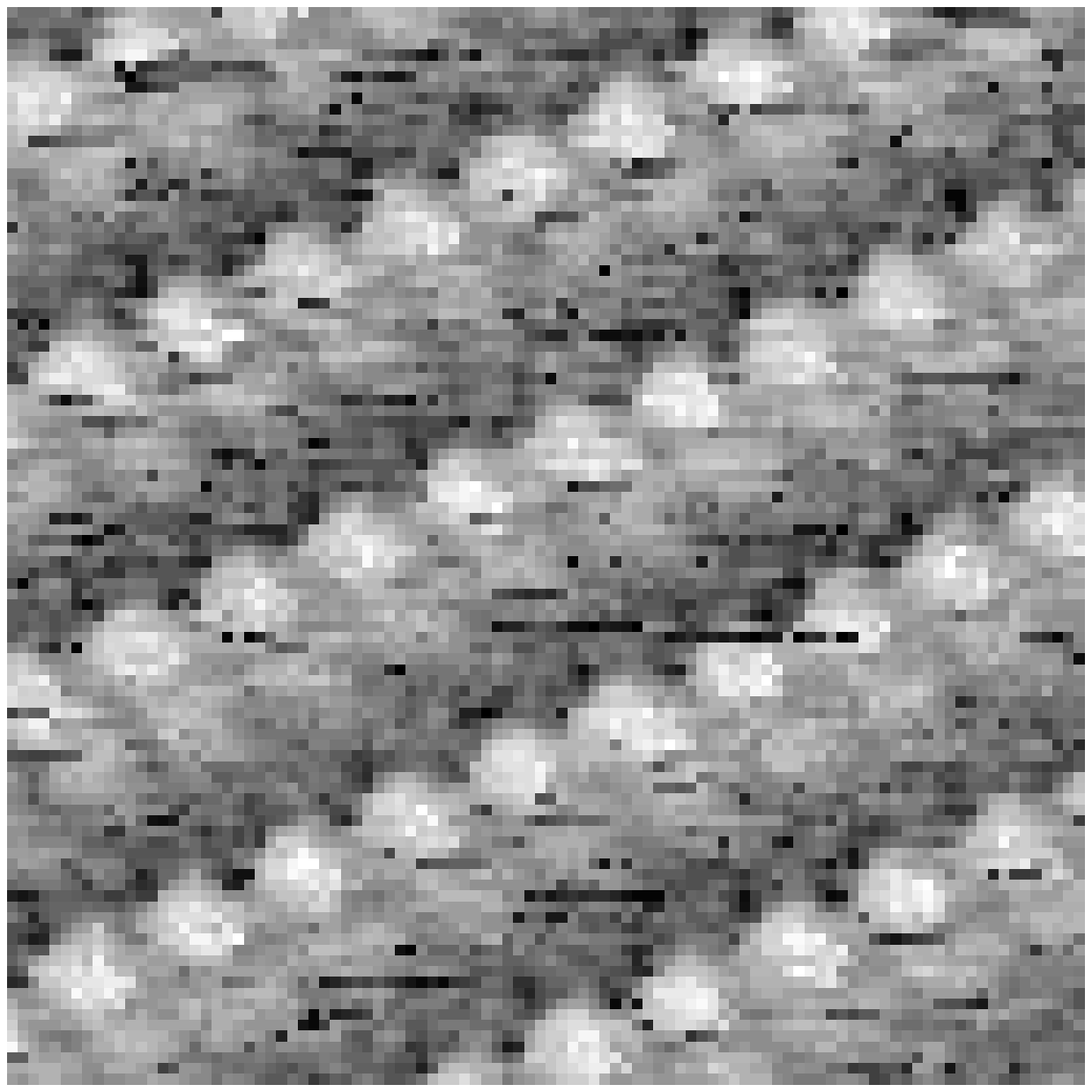} &
 \includegraphics[width=3.4cm]{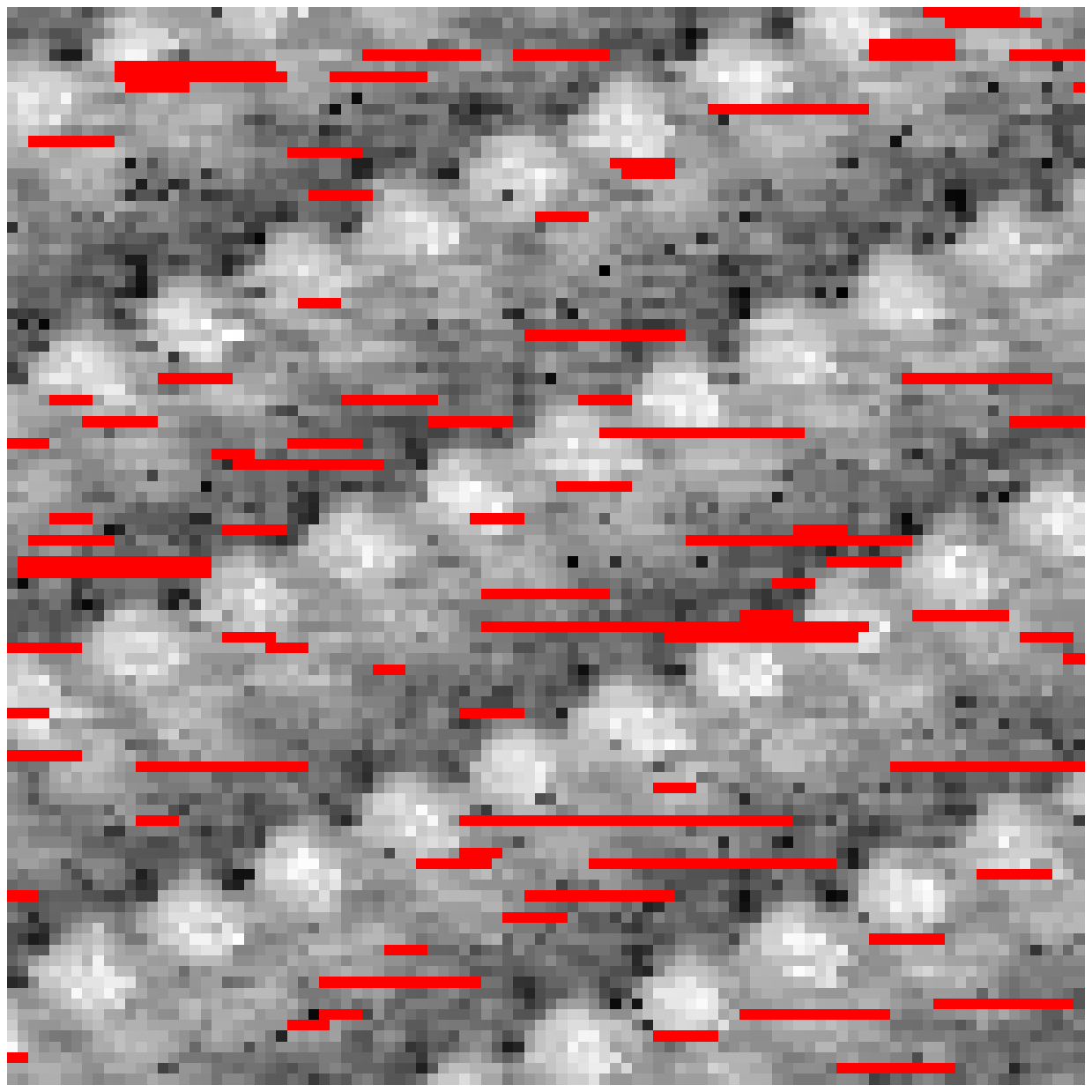} \\*[-0pt]
  (a) & (b)\\*[4pt]
 \includegraphics[width=3.4cm]{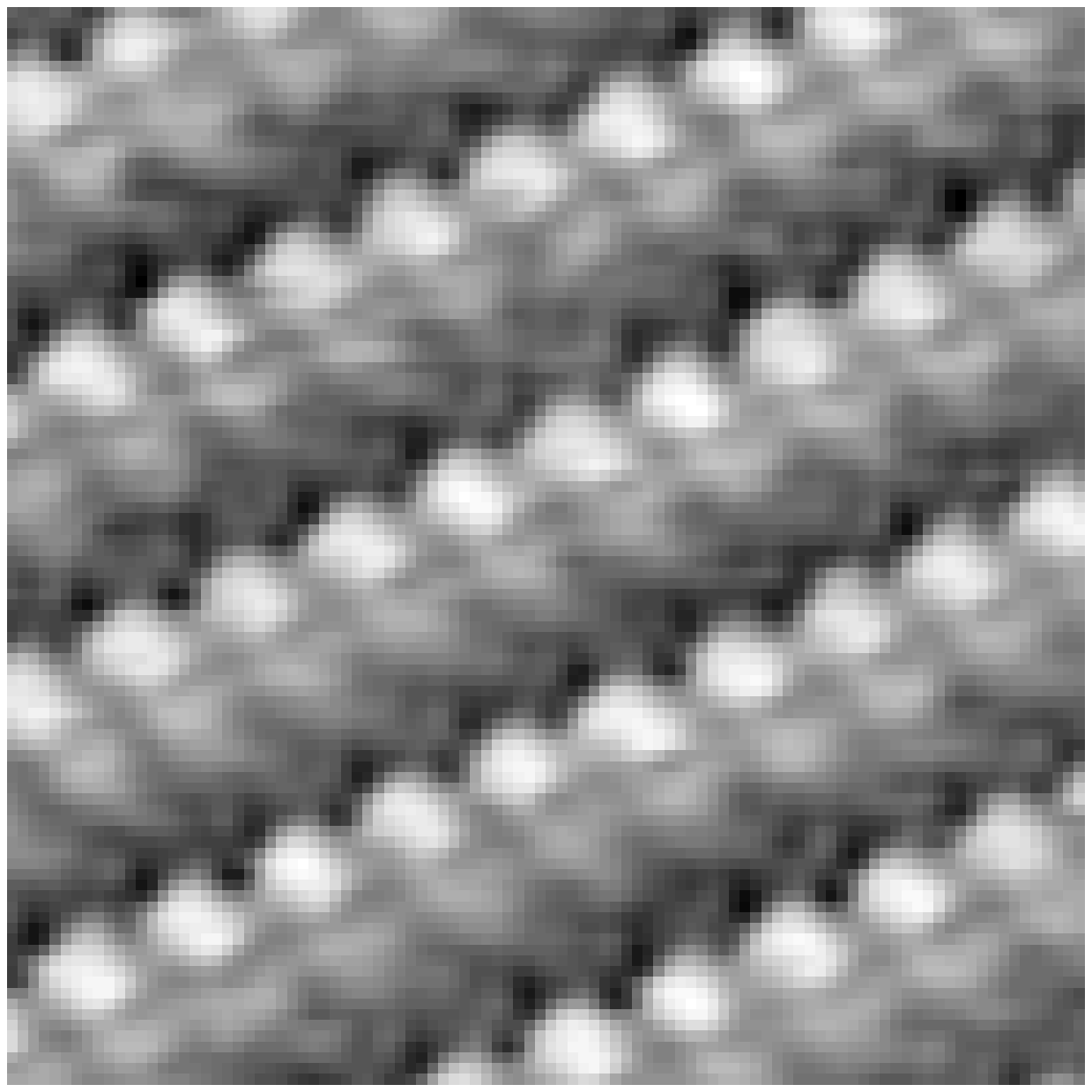} &
 \includegraphics[width=3.4cm]{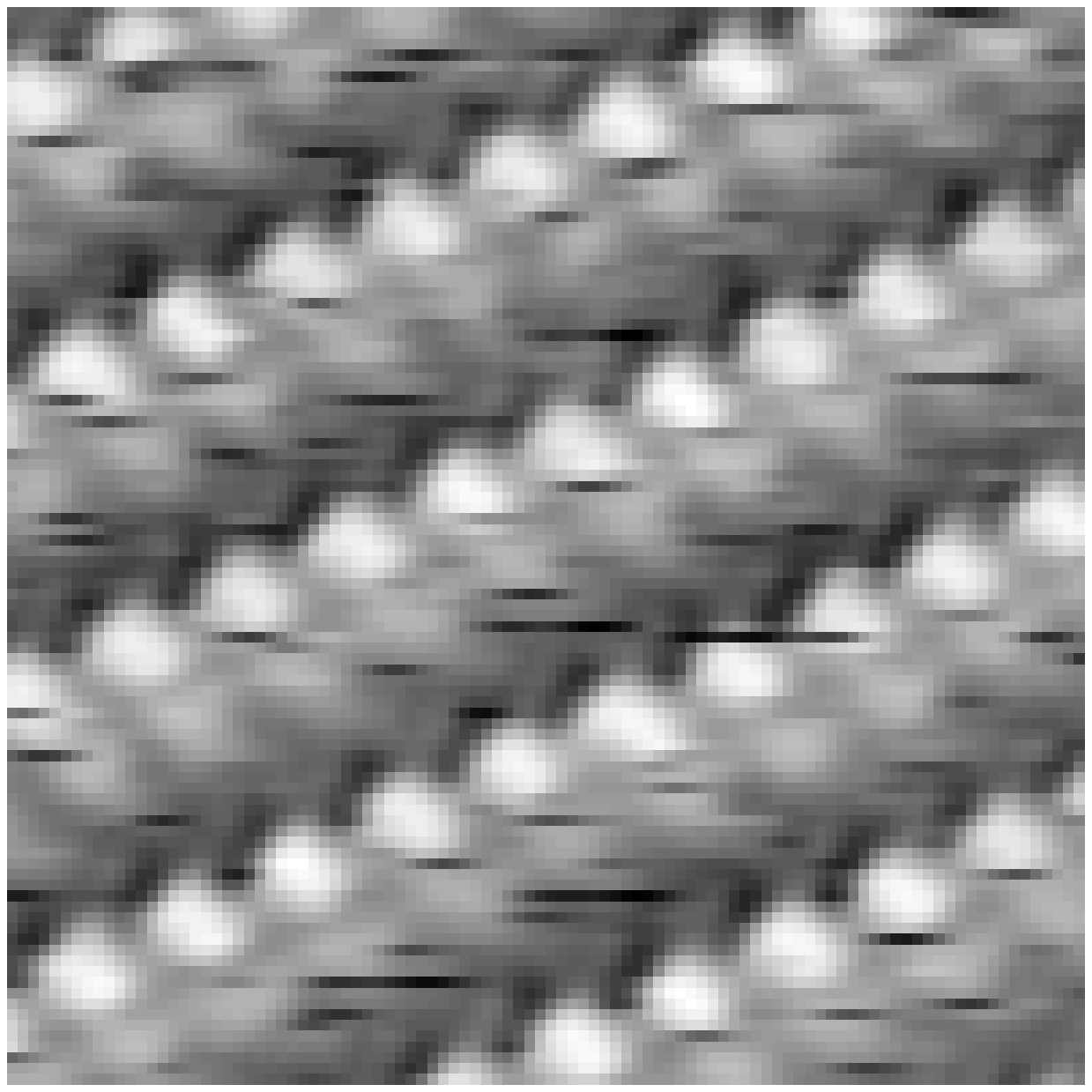} \\*[-0pt] 
 (c) & (d)\\*[4pt]
\end{tabular}
 \caption{Close up of STM image (30$\times$30nm, V=570mV, I=28pA) of a polymorphic phase of a ZnOEP monolayer absorbed on  HOPG where the ZnOEP are organized as dimers with different brightness: (a) noisy image; (b) noisy image with mask (in red); (c) restored with proposed algorithm; (d) restored with proposed algorithm without mask.\label{fig:results2c}}%
\end{figure}

To compare the inpainting results, we also run a version of the proposed algorithm without the mask.
Close ups of the noisy image, mask and restored images are depicted in Figure~\ref{fig:results2c}.

\begin{figure}
\begin{tabular}{c}
\includegraphics[width=7.3cm]{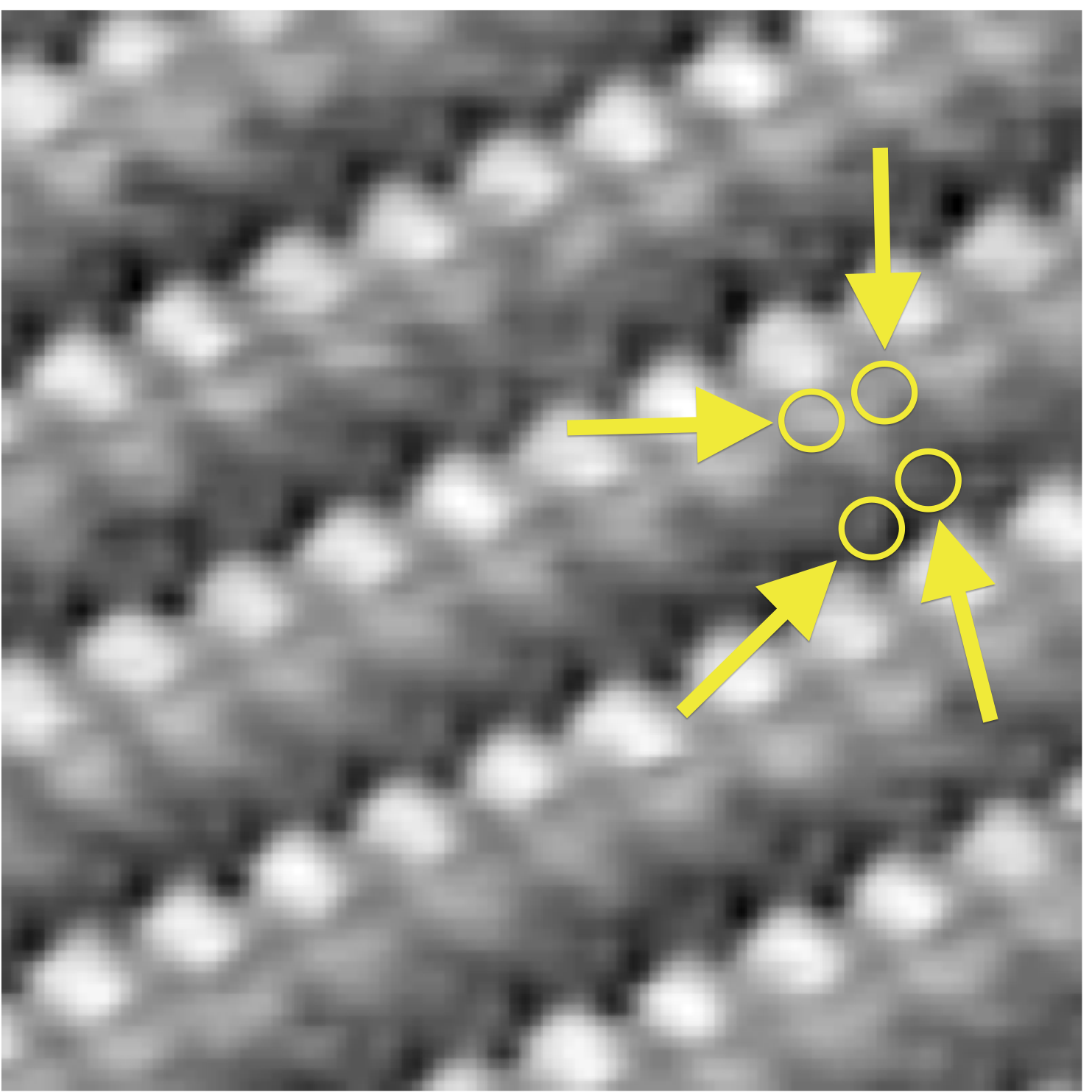}\\*[-0pt]
(a)\\*[4pt]
\includegraphics[width=8.5cm]{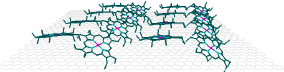}\\*[-0pt]
(b)\\*[4pt]
\end{tabular}
 \caption{(a) Detail of ZnOEP central part surrounded by 4 protrusions; and (b) monolayer molecular schema where the ZnOEP molecules that are parallel to the HOPG correspond to the brighter domains and the tilted ones correspond to the darker domains.\label{fig:results2d}}%
\end{figure}

In Figure~\ref{fig:results2c}~(c), the restored image (denoised) shows clearly the details of darker ZnOEP where it is possible to observe the ZnOEP central part surrounded by 4 protrusions that we attribute to the ethyl groups that are oriented towards the HOPG as is represented by molecular schema of Figure~\ref{fig:results2d}.

When the proposed algorithm is used without a mask, Figure~\ref{fig:results2c}~(d), the structured outliers are only partially removed. This happens because the dictionary learning algorithm actually learns these structures. Thus, there will be atoms available in $\mathbf{D}$ such that a patch with outliers still admits a sparse representation. However, when a mask is used, the algorithm removes these outliers and fill in the pixels with new values. Instead of a local median or a more sophisticated interpolation algorithm, by relying on the learned dictionary,  the filled in values are in accordance to the global image statistics.

\section{Conclusion}

We presented a denoising algorithm, formulated as sparse regression problem, to mitigate the effect of the degradation mechanisms observed in STM images. 
The proposed formulation was introduced based on the fact that STM images present a high level of self-similarity, so that a suitable dictionary could be learned from the noisy image. The algorithm also cope with the existence of artifacts, mainly dropouts, by treating them as missing data. The results obtained outperform those algorithms that substitute the outliers by a local filtering, as the filled in values are now in accordance to the global image statistics.

\section*{acknowledgement}
The authors thank FCT--Portugal for financial support under the project PEst--OE/EEI/LA0008/2013 and Post-Doc grant to QF.

\bibliographystyle{IEEEtranS}


\end{document}